\documentclass[10pt,twocolumn,letterpaper]{article}

\usepackage{iccv}

\usepackage[utf8]{inputenc}
\makeatletter
\@namedef{ver@everyshi.sty}{}
\makeatother
\usepackage{pgfplots}
\DeclareUnicodeCharacter{2212}{−}
\usepgfplotslibrary{groupplots,dateplot}
\usepackage{tikz}
\usetikzlibrary{patterns,shapes.arrows,positioning,fit}
\pgfplotsset{compat=newest}
\usepackage{ifthen}
\usepackage[most]{tcolorbox}

\usepackage{times}
\usepackage{epsfig}
\usepackage{graphicx}
\usepackage{amsmath}
\usepackage{amssymb}

\usepackage{array}
\usepackage{booktabs}
\usepackage{multirow}
\usepackage{multicol}
\usepackage{xcolor}
\usepackage{calrsfs}
\usepackage{caption}
\usepackage{subcaption}
\usepackage{wrapfig}

\DeclareMathAlphabet{\pazocal}{OMS}{zplm}{m}{n}

\usepackage[pagebackref=true,breaklinks=true,letterpaper=true,colorlinks,bookmarks=false]{hyperref}

\iccvfinalcopy 

\ificcvfinal\fi

\begin{document}

\title{Bit-Mixer: Mixed-precision networks with runtime bit-width selection}

\author{Adrian Bulat\\
Samsung AI Cambridge\\
{\tt\small adrian@adrianbulat.com}
\and
Georgios Tzimiropoulos\\
Samsung AI Cambridge\\
Queen Mary University of London \\
{\tt\small g.tzimiropoulos@qmul.ac.uk}
}

\maketitle
\ificcvfinal\thispagestyle{empty}\fi

\begin{abstract}

Mixed-precision networks allow for a variable bit-width quantization for every layer in the network. A major limitation of existing work is that the bit-width for each layer must be predefined during training time. This allows little flexibility if the characteristics of the device on which the network is deployed change during runtime. In this work, we propose Bit-Mixer, the very first method to train a meta-quantized network where during test time any layer can change its bid-width without affecting at all the overall network's ability for highly accurate inference. To this end, we make 2 key contributions: (a) Transitional Batch-Norms, and (b) a 3-stage optimization process which is shown capable of training such a network. We show that our method can result in mixed precision networks that exhibit the desirable flexibility properties for on-device deployment without compromising accuracy. Code will be made available.

\end{abstract}

\section{Introduction}

Deep Neural Networks have reached state-of-the-art accuracy across a plethora of computer vision and machine learning tasks. Despite their unprecedented accuracy, directly deploying such models on devices with limited computational resources and/or power constraints remains prohibitive. To address this problem, a series of related research directions have emerged such as network pruning~\cite{liu2018rethinking,molchanov2019importance,li2016pruning}, network compression~\cite{lebedev2014speeding,ullrich2017soft,li2019learning}, neural architecture search~\cite{liu2018darts,chen2019progressive} and network quantization. The later offers the most straightforward improvements as using fewer bits for the weights and activations significantly reduces the compute and storage requirements. For example, switching from FP32 to Int-8 precision, a $4 \times$ improvement in terms of speed and storage is obtained without any bells and whistles. This paper is on mixed-precision networks which allow for a variable bit-width quantization for every layer in the network. 

Mixed-bit precision networks allow for a finer granularity of  quantization at a layer level and, hence, offer practical advantages in terms of finding a more optimal trade-off between efficiency (i.e. speed) and memory requirements, and network accuracy. While this is more flexible than having the same bit-width across the whole network, mixed-bit precision approaches have also their own limitations. Firstly, due to an ever-growing number of different hardware platforms that a developer needs to support, each with its own unique characteristics and capabilities, quantizing networks, partially or fully, with mixed-bit precision in order to obtain an optimal trade-off between accuracy and speed becomes challenging. Secondly, and more importantly, even on the same device, due to either other concurrent processes running, battery level, temperature or simply prioritization, the available resources can vary. Ideally, a network should be able to dynamically react to these changes and adapt its quantization level per layer or module \textit{on the fly} without incurring undesirable, or even more importantly, unpredictable penalties on inference accuracy. 

The method we propose in this paper, coined \textbf{Bit-Mixer}, attempts to provide an answer to the aforementioned challenges. Bit-Mixer shifts away the focus from finding the optimal bid-with allocation per layer during training as done in \textit{all previous} work. Instead, we propose to train a meta-quantized network which during test time can switch to \text{any quantization level} for \text{any layer} in the network. Training such meta-networks is however non-trivial due to the exponential number of unique combinations, the weight sharing constraint across different bit-widths, and the drastic variations in representational power that occur when the bit-width changes (e.g. 4 bits vs 1 bit). To this end, we make the following \textbf{contributions}:

\begin{figure*}[ht]
    \centering
        \begin{subfigure}[t]{0.3\textwidth}
            \centering
            \scalebox{0.7}{\input{figures/_shared_nbits}

\begin{tikzpicture}[x=1.5cm, y=1.5cm, >=stealth]

\foreach \m/\l [count=\x] in {1,2,missing,3} 
  {
      \ifthenelse{\x=3}
      {
         \node [every neuron/.try, neuron \m/.try, fill=none] (input-\m) at (\x,0) {};
      }
      {
        \pgfmathparse{\colorPallete[\x-1]};
        \definecolor{currentColor}{rgb}{\pgfmathresult};
         \node [every neuron/.try, neuron \m/.try, fill=currentColor] (input-\m) at (\x,0) {};
      }
     
  }
  
  \foreach \m [count=\x] in {1,2,missing,3}
  {
      \ifthenelse{\x=3}
      {
          \node [every neuron/.try, neuron \m/.try ] (hidden-\m) at (\x,-1.5) {};
      }
      {
        \pgfmathparse{\colorPallete[\x-1]};
        \definecolor{currentColor}{rgb}{\pgfmathresult};
         \node [every neuron/.try, neuron \m/.try, fill=currentColor] (hidden-\m) at (\x,-1.5) {};
      }
  }
  
    \foreach \m [count=\x] in {1,2,missing,3}
  {
      \ifthenelse{\x=3}
      {
          \node [every neuron/.try, neuron \m/.try ] (output-\m) at (\x,-3) {};
      }
      {
        \pgfmathparse{\colorPallete[\x-1]};
        \definecolor{currentColor}{rgb}{\pgfmathresult};
         \node [every neuron/.try, neuron \m/.try, fill=currentColor] (output-\m) at (\x,-3) {};
      }
  }
  
\foreach \l [count=\i] in {1,2,n}
  \draw [<-] (input-\i) -- ++(0,1)
    node [left, midway] {$I_\l$};


\foreach \l [count=\i] in {1,2,n}
  \draw [->] (output-\i) -- ++(0,-1)
    node [left, midway] {$O_\l$};

\foreach \i in {1,...,3}
  \foreach \j in {1,...,3}
   {
        \ifthenelse{\i=\j} {
             \draw [->] (input-\i) -- (hidden-\j);
        }
    }

\foreach \i in {1,...,3}
  \foreach \j in {1,...,3}
    {
        \ifthenelse{\i=\j} {
            \draw [->] (hidden-\i) -- (output-\j);
        }
    }


\end{tikzpicture}}
            \caption{\textbf{Independent:} Each bit-width requires training a new network with independent weights.}
            \label{fig:q-independent}
    \end{subfigure}
    ~
    \begin{subfigure}[t]{0.3\textwidth}
            \centering
            \scalebox{0.7}{\input{figures/_shared_nbits}

\begin{tikzpicture}[x=1.5cm, y=1.5cm, >=stealth]

\foreach \m/\l [count=\x] in {1,2,missing,3} 
  {
      \ifthenelse{\x=3}
      {
         \node [every neuron/.try, neuron \m/.try, fill=none] (input-\m) at (\x,0) {};
      }
      {
        \pgfmathparse{\colorPallete[\x-1]};
        \definecolor{currentColor}{rgb}{\pgfmathresult};
         \node [every neuron/.try, neuron \m/.try, fill=currentColor] (input-\m) at (\x,0) {};
      }
     
  }
  
  \foreach \m [count=\x] in {1,2,missing,3}
  {
      \ifthenelse{\x=3}
      {
          \node [every neuron/.try, neuron \m/.try ] (hidden-\m) at (\x,-1.5) {};
      }
      {
        \pgfmathparse{\colorPallete[\x-1]};
        \definecolor{currentColor}{rgb}{\pgfmathresult};
         \node [every neuron/.try, neuron \m/.try, fill=currentColor] (hidden-\m) at (\x,-1.5) {};
      }
  }
  
    \foreach \m [count=\x] in {1,2,missing,3}
  {
      \ifthenelse{\x=3}
      {
          \node [every neuron/.try, neuron \m/.try ] (output-\m) at (\x,-3) {};
      }
      {
        \pgfmathparse{\colorPallete[\x-1]};
        \definecolor{currentColor}{rgb}{\pgfmathresult};
         \node [every neuron/.try, neuron \m/.try, fill=currentColor] (output-\m) at (\x,-3) {};
      }
  }
  
  \node[draw,dotted,fit=(input-1) (input-2) (input-3),line width=1] (D1) {} node[right = 0cm of D1] {$W_1$};
  \node[draw,dotted,fit=(hidden-1) (hidden-2) (hidden-3),line width=1] (D2) {} node[right = 0cm of D2] {$W_l$};
  \node[draw,dotted,fit=(output-1) (output-2) (output-3),line width=1] (D3) {} node[right = 0cm of D3] {$W_L$};
  
\foreach \l [count=\i] in {1,2,n}
  \draw [<-] (input-\i) -- ++(0,1)
    node [left, midway] {$I_\l$};


\foreach \l [count=\i] in {1,2,n}
  \draw [->] (output-\i) -- ++(0,-1)
    node [left, midway] {$O_\l$};

\foreach \i in {1,...,3}
  \foreach \j in {1,...,3}
   {
        \ifthenelse{\i=\j} {
             \draw [->] (input-\i) -- (hidden-\j);
        }
    }

\foreach \i in {1,...,3}
  \foreach \j in {1,...,3}
    {
        \ifthenelse{\i=\j} {
            \draw [->] (hidden-\i) -- (output-\j);
        }
    }


\end{tikzpicture}}
            \caption{\textbf{Adabits:} A single network can be quantized to any of $n$ bit-widths at runtime. All layers inside the network share the \textit{same} bit-width.}
            \label{fig:q-adabits}
    \end{subfigure}
    ~
    \begin{subfigure}[t]{0.3\textwidth}
            \centering
            \scalebox{0.7}{\input{figures/_shared_nbits}

\begin{tikzpicture}[x=1.5cm, y=1.5cm, >=stealth]

\foreach \m/\l [count=\x] in {1,2,missing,3} 
  {
      \ifthenelse{\x=3}
      {
         \node [every neuron/.try, neuron \m/.try, fill=none] (input-\m) at (\x,0) {};
      }
      {
        \pgfmathparse{\colorPallete[\x-1]};
        \definecolor{currentColor}{rgb}{\pgfmathresult};
         \node [every neuron/.try, neuron \m/.try, fill=currentColor] (input-\m) at (\x,0) {};
      }
     
  }
  
  \foreach \m [count=\x] in {1,2,missing,3}
  {
      \ifthenelse{\x=3}
      {
          \node [every neuron/.try, neuron \m/.try ] (hidden-\m) at (\x,-1.5) {};
      }
      {
        \pgfmathparse{\colorPallete[\x-1]};
        \definecolor{currentColor}{rgb}{\pgfmathresult};
         \node [every neuron/.try, neuron \m/.try, fill=currentColor] (hidden-\m) at (\x,-1.5) {};
      }
  }
  
    \foreach \m [count=\x] in {1,2,missing,3}
  {
      \ifthenelse{\x=3}
      {
          \node [every neuron/.try, neuron \m/.try ] (output-\m) at (\x,-3) {};
      }
      {
        \pgfmathparse{\colorPallete[\x-1]};
        \definecolor{currentColor}{rgb}{\pgfmathresult};
         \node [every neuron/.try, neuron \m/.try, fill=currentColor] (output-\m) at (\x,-3) {};
      }
  }
  
  \node[draw,dotted,fit=(input-1) (input-2) (input-3),line width=1] (D1) {} node[right = 0cm of D1] {$W_1$};
  \node[draw,dotted,fit=(hidden-1) (hidden-2) (hidden-3),line width=1] (D2) {} node[right = 0cm of D2] {$W_l$};
  \node[draw,dotted,fit=(output-1) (output-2) (output-3),line width=1] (D3) {} node[right = 0cm of D3] {$W_L$};

\foreach \l [count=\i] in {1,2,n}
  \draw [<-] (input-\i) -- ++(0,1)
    node [left, midway] {$I_\l$};


\foreach \l [count=\i] in {1,2,n}
  \draw [->] (output-\i) -- ++(0,-1)
    node [left, midway] {$O_\l$};

\foreach \i in {1,...,3}
  \foreach \j in {1,...,3}
    \draw [->] (input-\i) -- (hidden-\j);

\foreach \i in {1,...,3}
  \foreach \j in {1,...,3}
    \draw [->] (hidden-\i) -- (output-\j);


\end{tikzpicture}}
            \caption{\textbf{Proposed method (Bit-Mixer):} A single network whose individual layers can be quantized at runtime to any bit-width, without any re-training, resulting in an exponential number of mixed precision networks that one can choose from to fit the device characteristics and computational resources available on-the-fly.}
            \label{fig:q-ours}
    \end{subfigure}
    \caption{Comparison between prior network quantization paradigms (a,b) and ours (c).}
    \label{fig:my_label}
\end{figure*}

\begin{enumerate}
    \item 
    Transitional Batch-Norms: To properly compensate for the distribution shift that arises when a change in the bit-width occurs between two consecutive layers, for each transition between different bit-widths, we propose to learn a separate batch normalization layer, coined Transitional Batch-Norm.
    \item
    3-stage Optimization: We firstly propose an efficient 2-stage process to train an intermediate meta-network which at runtime can select different bit-widths which however are shared across the entire network. Then, a 3-rd final stage is introduced to gradually transition from the intermediate meta-network to the final one where the quantization level can be randomly selected at a block or layer level. Notably,
    our meta-network uses a single, shared set of weights.
    \item We conducted a number of ablation studies which shed light into the behaviour of several components of our method. Moreover, building on top of the findings of ~\cite{dong2019hawq}, we analyze Bit-Mixer's sub-nets exploring the inter-dependencies between the accuracy and the quantization level selected for a given layer. Finally, we extensively evaluated the accuracy of the proposed Bit-Mixer across different architectures and model sizes. 
\end{enumerate}

\section{Related work}

Network quantization aims to alleviate the high computational and memory cost of modern deep neural networks by using fewer bits (\ie $b<32$) for the weights and activations. Most of early works quantized the weights only~\cite{han2015deep,courbariaux2015binaryconnect}. Follow-up works quantize both the weights and the activations while maintaining the same bit-width across the entire network using uniform quantization schemes~\cite{jacob2018quantization,jin2019towards,zhang2018lq,mellempudi2017ternary,rastegari2016xnor,bulat2019xnor,zhou2016dorefa,zhuang2020training,esser2020learned}. 

More recently, a growing body of work explores mixed-precision quantization which enables, within the same architecture, different layers to use different bit-widths~\cite{elthakeb2019releq,uhlich2019differentiable,wang2019haq}. The bit-width allocation process is typically performed either using reinforcement learning techniques~\cite{elthakeb2019releq,wang2019haq} or differential  search~\cite{wu2018mixed,uhlich2019differentiable}. Contrary to Bit-Mixer (our work), all the aforementioned methods result in a single network with different but pre-defined bit-widths per layer that cannot be modified without retraining.

Related to our work is the line of research somewhat related to Neural Architecture Search (e.g. ~\cite{chen2019progressive,stamoulis2019single,liu2018darts}), and, in particular, the works of~\cite{yu2018slimmable,cai2019once} where the authors train a super-network from which sub-nets with varying depth, width and kernel size can be sampled without retraining. These works do not consider the problem of network quantization at all.

More closely related to our work is AdaBits~\cite{jin2020adabits} where the authors propose to train a single neural network, with a shared set of weights, that can switch bit-width at runtime. However, a major limitation of AdaBits is that it is not a mixed-precision network: it uses the same bit-width across the entire network which reduces its flexibility in practical scenarios. Moreover, from a methodological perspective, the Transitional Batch-Norms as well as the 3-stage optimization procedure proposed in our work are fundamentally different from the methods described in~\cite{jin2020adabits}. 

\begin{figure*}[!htbp]
    \centering
    \hspace{-4em}
    \begin{subfigure}[t]{0.3\textwidth}
        \includegraphics[width=6.5cm]{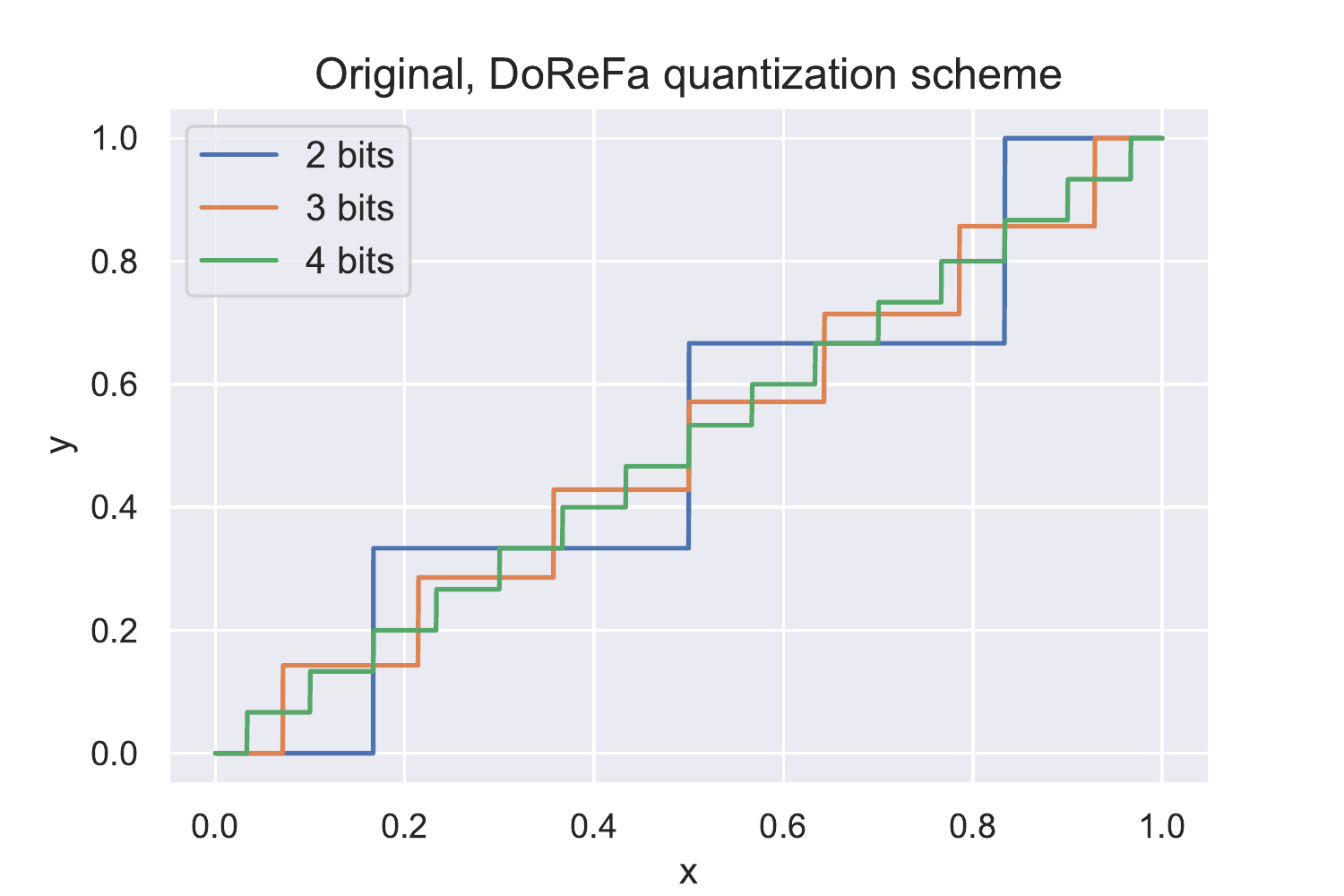}
    \end{subfigure}
    ~
    \begin{subfigure}[t]{0.3\textwidth}
        \includegraphics[width=6.5cm]{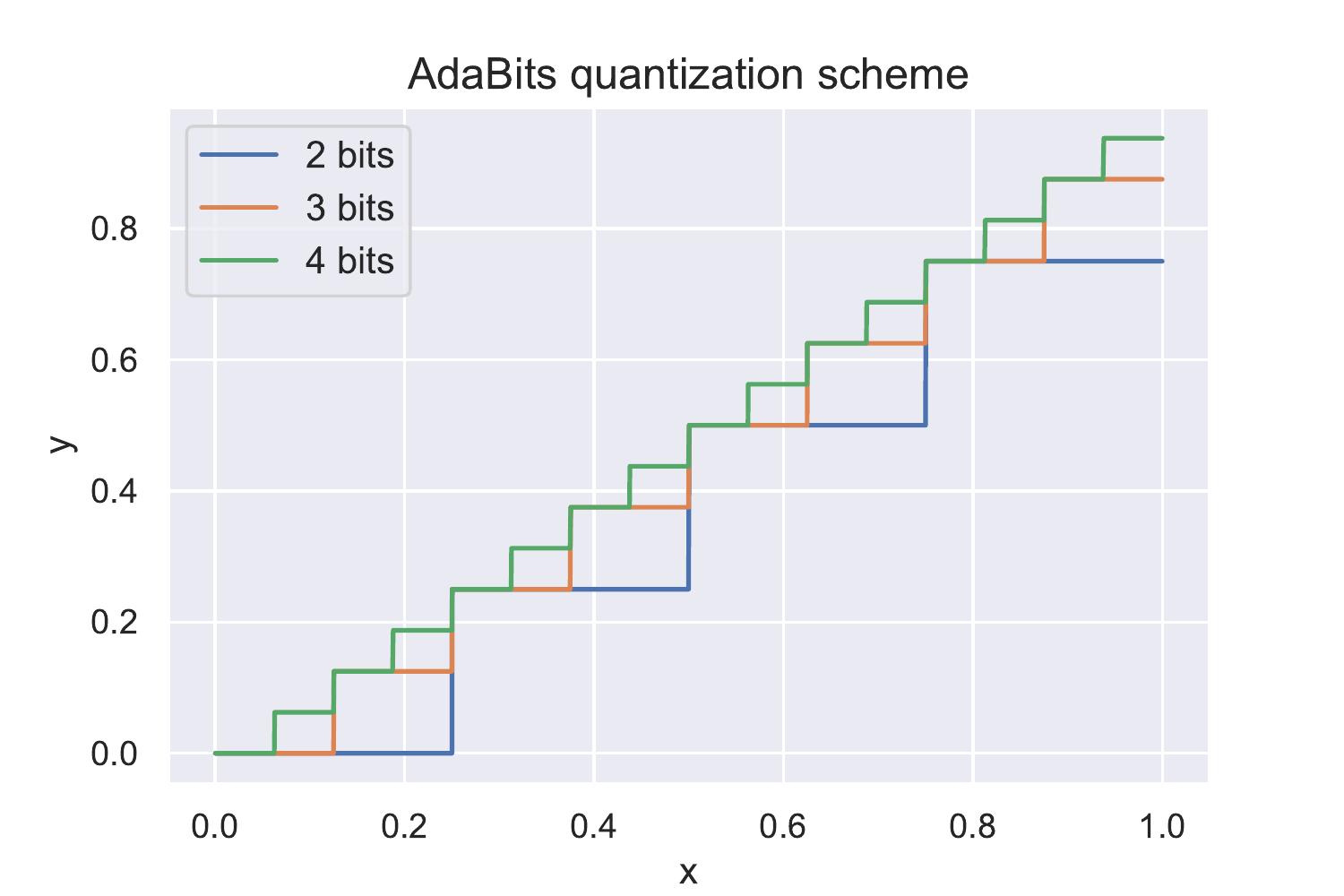}
    \end{subfigure}
    ~
    \begin{subfigure}[t]{0.3\textwidth}
        \includegraphics[width=6.5cm]{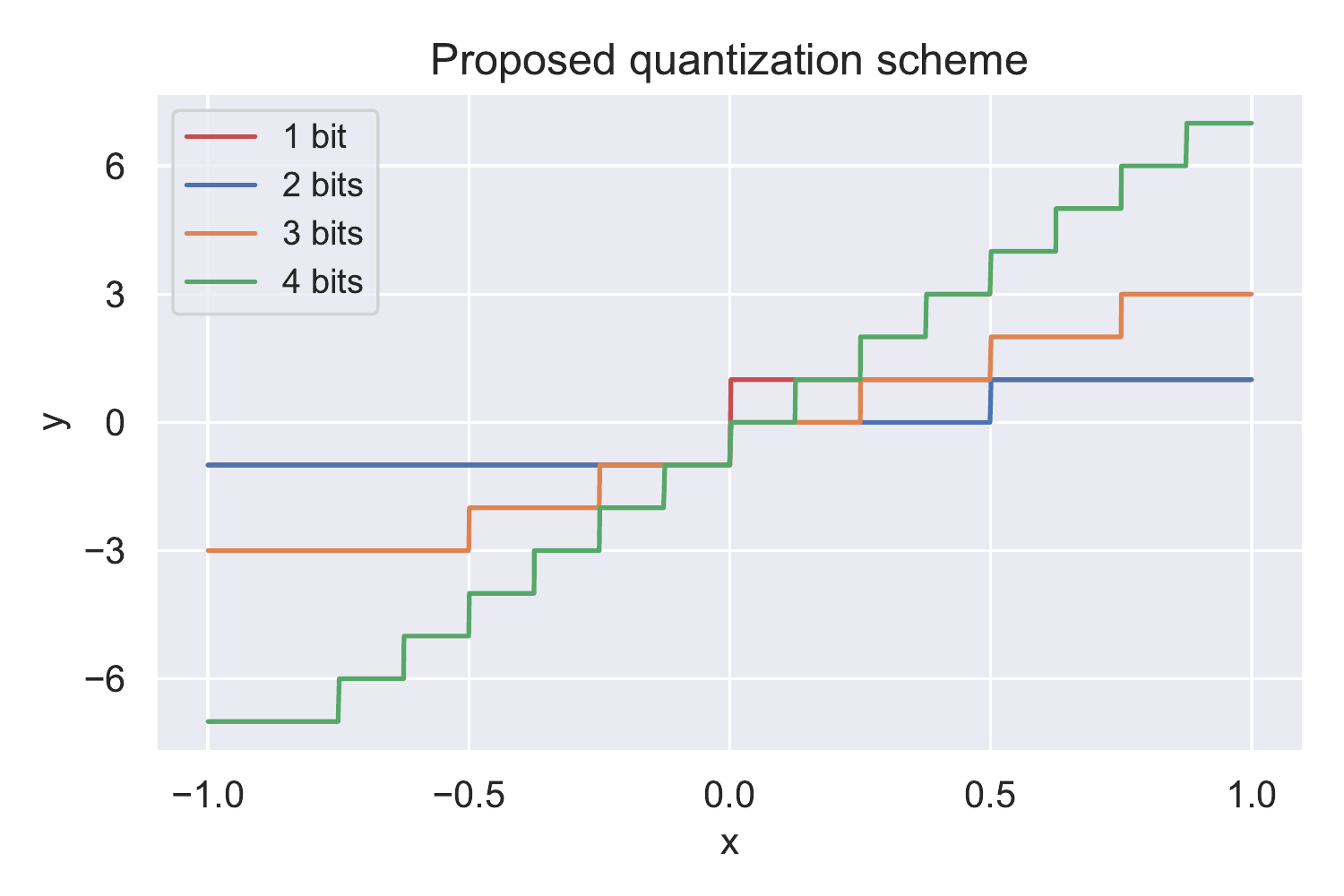}
    \end{subfigure}
    \caption{Difference between various quantization schemes (from left to right) used in DoReFa~\cite{zhou2016dorefa}, AdaBits~\cite{jin2020adabits} and Bit-Mixer (Ours). In all cases $y=\texttt{quant}(x)$}
    \label{fig:quantization-schemes}
\end{figure*}

\section{Method}\label{sec:method}

\subsection{Unifying 1--n bit quantization} 

For a given layer $l$,  we denote the quantization of the weights $\mathbf{W}$ and input activations $\mathbf{A}$  as $\texttt{quant}(\mathbf{W}, b) = \Tilde{\mathbf{W}}_b$ and $\texttt{quant}(\mathbf{A}, b) = \Tilde{\mathbf{A}}_b$, respectively, where $b=\{1,2,\dots,n\}$ denotes the bit-width.

For the quantization function, we opted to adopt and adapt the recently proposed LSQ~\cite{esser2020learned} as follows: to handle both cases $b=1$ (i.e. binary networks) and $1<b\leq n$, we quantize both the activations and the weights between $(-m_b, m_b)$, where $m_b = 2^{b-1}-1$ is the maximum value representable using $b$ bits \footnote{This is because in binary networks both the weights and the activations are quantized using the sign function~\cite{rastegari2016xnor}, hence a symmetric quantizer is needed.}. Although this symmetric quantization discards 1 state, we will show that this has no impact on the accuracy of the quantized networks. Furthermore the case $b=2$ will degenerate in what is known in literature as ternary quantization, allowing for further specific optimizations made possible by the induced sparsification~\cite{zhu2016trained}. Overall, our unified quantization scheme is defined as follows:
\begin{equation}
\label{eq:ours-q}
\begin{split}
\mathbf{\Tilde{W}}_b = \texttt{q}_b(W) \\ 
\mathbf{\Tilde{A}}_b = \texttt{q}_b(A),
\end{split}
\end{equation}
where the quantization function $\texttt{q}_b(x)$ is computed as:
\begin{equation}
    \label{eq:ours-q-op}
    \begin{split}
    \texttt{q}_b(x) = \alpha \times \texttt{q'}(\texttt{clip}(\frac{x}{\alpha}, -m_b, m_b)) \\
    \texttt{q'}(x) = \begin{cases} \lfloor.\rfloor, & \mbox{if } b>1 \\ \texttt{sign}, & \mbox{if } b=1 \end{cases},
    \end{split}
\end{equation}
where $\lfloor.\rfloor$ is the \texttt{floor} rounding operator. Notice that we replaced the \texttt{round} function used in LSQ~\cite{esser2020learned} with \texttt{floor}. This allows us to obtain the weights $\mathbf{\Tilde{W}}_i$ directly from $\mathbf{\hat{W}}_{i+1}$ without the need of storing the full precision weights, significantly reducing the model storage requirements (as its size is determined solely by the size of $\mathbf{\hat{W}}_n$). The difference between various quantization schemes used for mixed precision networks is shown in Fig. \ref{fig:quantization-schemes}.

\subsection{Transitional Batch-Norm}\label{ssec:transitional-bn}

Quantizing the individual layers and blocks to different bit-widths will result in features that follow different distributions. This is because of two reasons: Firstly, it is a consequence of the inherent change in the representational power due to the change of precision. Secondly, as the number of bits drops, the network is unable to approximate closely the feature distribution of higher bit-widths, as the weight distribution significantly changes (this can also be seen in Fig.~\ref{fig:weights-distribution} for $b=\{1,2,3,4\}$).

To properly compensate for the distribution shift that arises when a change in the bit-width occurs between two consecutive layers, for each transition between different bit-widths, we propose to learn a separate batch normalization layer, coined Transitional Batch-Norm. Specifically, if $1\leq i \leq n$ is the bit-with of layer $l-1$ and $1\leq j \leq n$ is the bit-with of layer $l$, we learn BN parameters $\alpha_{ij}$ and $\beta_{ij}$.
The parameters $\alpha_{ij}$ and $\beta_{ij}$ remain tied to the bit-width $j$ of the layer $l$ since they depend on the current quantization level alone, irrespectively of the layer's weights, which do not undergo a transition as opposed to the activations. We note that introducing the Transitional Batch-Norm layers does not induce any increase in the complexity of the network; only a small increase in network size is introduced (less than 1\% of the total parameters count). Importantly, we emphasize that, without the Transitional Batch-Norms, the network is unable to converge to a satisfactory level of accuracy. This phenomenon is present both when training from scratch and when initializing from a pretrained model (see also Table~\ref{tab:resnet18_transit_bn}).

\subsection{Optimization process}\label{ssec:optimization-strategy}

A key remaining aspect of our method is how to train the proposed meta-network which turns out to be very challenging for several reasons. A direct naive approach, where all the paths are active simultaneously, is unfeasible due to both memory and computational constraints. Besides this, we considered an approximation to this training where all active paths are considered between 2 adjacent layer. Even in this case, we found the models unstable to train due to the internal competition  arising, especially early in the training. 

A more computationally feasible approach is to select randomly (with equal probability) during training an active sub-path or a set of active sub-paths. However, in our experiments, we found that this leads to networks in which the accuracy of all bit-widths are closely tight together, pulling them towards the one with the lowest accuracy, and, hence, largely diminishing the potential advantages of training the proposed meta-network.

In order to successfully train the newly introduced quantized meta-network, we firstly propose an \textit{efficient} way to train a meta-network with can work at runtime with different bit-widths which however shared across \textit{the entire} network (Stages I \& II below). Then, to obtain the final meta-network, we propose to \textit{progressively} train the previous network by gradually transitioning from networks where all the layers are quantized to the same bit-width to ones where the quantization level is randomly selected \textit{at a block or layer} level (Stage III below). The procedure can be summarized as follows:

\noindent \textbf{Stage I:} During this stage, the network weights are kept real-valued while the activations (\ie features) are quantized to $n$ different bit-widths. Specifically, at each iteration, we randomly select, with equal probability, a bit-width $b$ out of the predefined set $\{1,\dots,n\}$. At this stage, the model will use the \textit{same bit-width} for the activations across all layers of the network. 
    
\noindent \textbf{Stage II:} During this stage, we use the network trained in Stage I as initialization and repeat the process of the previous stage, with the difference being that this time both the weights and the activations are quantized. Again, the model will use the \textit{same bit-width} for both weights and activations across all layers of the network. Note that Adabits~\cite{jin2020adabits} trains a network similar to the one obtained at the end of this stage. Compared to Adabits which requires $n$ training stages, our scheme is more efficient requiring only 2 stages independently of $n$.
    
\noindent \textbf{Stage III:} Continuing the training process by resuming from the previous checkpoint, during this stage, and with probability $\sigma$, the weights and features are trained in the same fashion as described in Stage II (\ie the same bit-width is used across all layers). For the rest of time, \ie. with probability $1-\sigma$, the bit-width $b$ of each individual layer is randomly selected \textit{independently} of each other, resulting in a network where different bit-widths are used for different layers. As the training progresses, we gradually decrease $\sigma$, effectively increasing the chance of training the meta-network with layer-wise random bit-with allocation. We continue the process until $1-\sigma = k$, where $k$ is typically $\frac{3}{4}$. All 3 stages share the same training scheduler.
\section{Ablation studies}

Unless otherwise stated, we conduct our ablation studies by using Bit-Mixer to train a meta-ResNet-18~\cite{he2016deep} on ImageNet. We mainly report the accuracy of Bit-Mixer for the following 2 cases (note there is only one single network that is evaluated): \text{fixed} bit-width selection across all layers and \text{random} bit-width selection for each individual layer. For the latter case, we simply randomize the layer-wise bit-width selection for every iteration (forward pass) of the validation set. We note that this random layer-wise bit-width selection has been \textit{intentionally} chosen for Bit-Mixer's evaluation protocol since it conclusively shows that Bit-Mixer works as expected. However, in Section~\ref{ssec:hawq}, we \textit{do provide} accuracy results for the case where a simple method, based on~\cite{dong2019hawq}, has been used in order to discover high performing sub-nets within the trained meta-network.

\subsection{Effect of Transitional Batch-Norm}

\begin{figure}
    \centering
    \includegraphics[width=7.0cm]{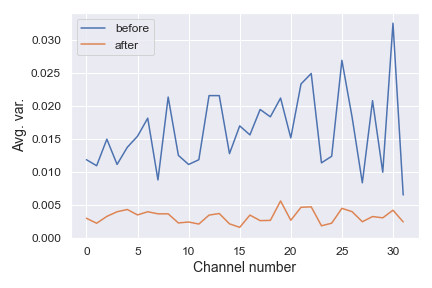}
    \caption{Variance per channel over the quantized activations $\Tilde{\mathbf{A}}_b, b=2,3,4$ before and after applying the Transitional Batch-Norm. Notice that the layer helps reducing the variance of the quantized activations significantly, stabilizing the training of the Bit-Mixer meta-network.}
    \label{fig:transit-bn}
\end{figure}

In Section~\ref{ssec:transitional-bn}, we introduced the Transitional Batch-Norm layers to compensate for the distribution shift between adjacent layers that are quantized to different bit-widths. Herein, we show their importance in terms of effectively training Bit-Mixer. As the results from Table~\ref{tab:resnet18_transit_bn} show, without Transitional Batch-Norm, the meta-network is unable to converge to a good solution although it was initialized using a model trained up to Stage II. The effect can be also observed by analyzing the statistics of the the features before and after applying the transitional batch norm layer in Fig.~\ref{fig:transit-bn}.

\begin{table}[!htbp]
	\caption{Top-1 accuracy (\%) on ImageNet for Bit-Mixer trained with and without Transitional BN.}\label{tab:resnet18_transit_bn}
	\centering
		\begin{tabular}{ccccc}
			\toprule

			\multirow{2}{*}{Bit-Mixer}  &\multicolumn{4}{c}{Bit-width} \\
			\cline{2-5} 
			& 4 & 3 & 2 & Rand.  \\			
            \midrule
            w/o Transitional BN & 8.2 & 5.6 & 10.2  & 8.8 \\
            with Transitional BN  & 69.2 & 68.6 & 64.4 & 65.8 \\
			\bottomrule
		\end{tabular}
\end{table}

\subsection{Analyzing Bit-Mixer's sub-nets}\label{ssec:hawq}

\begin{figure}
    \centering
    \includegraphics[width=7.0cm]{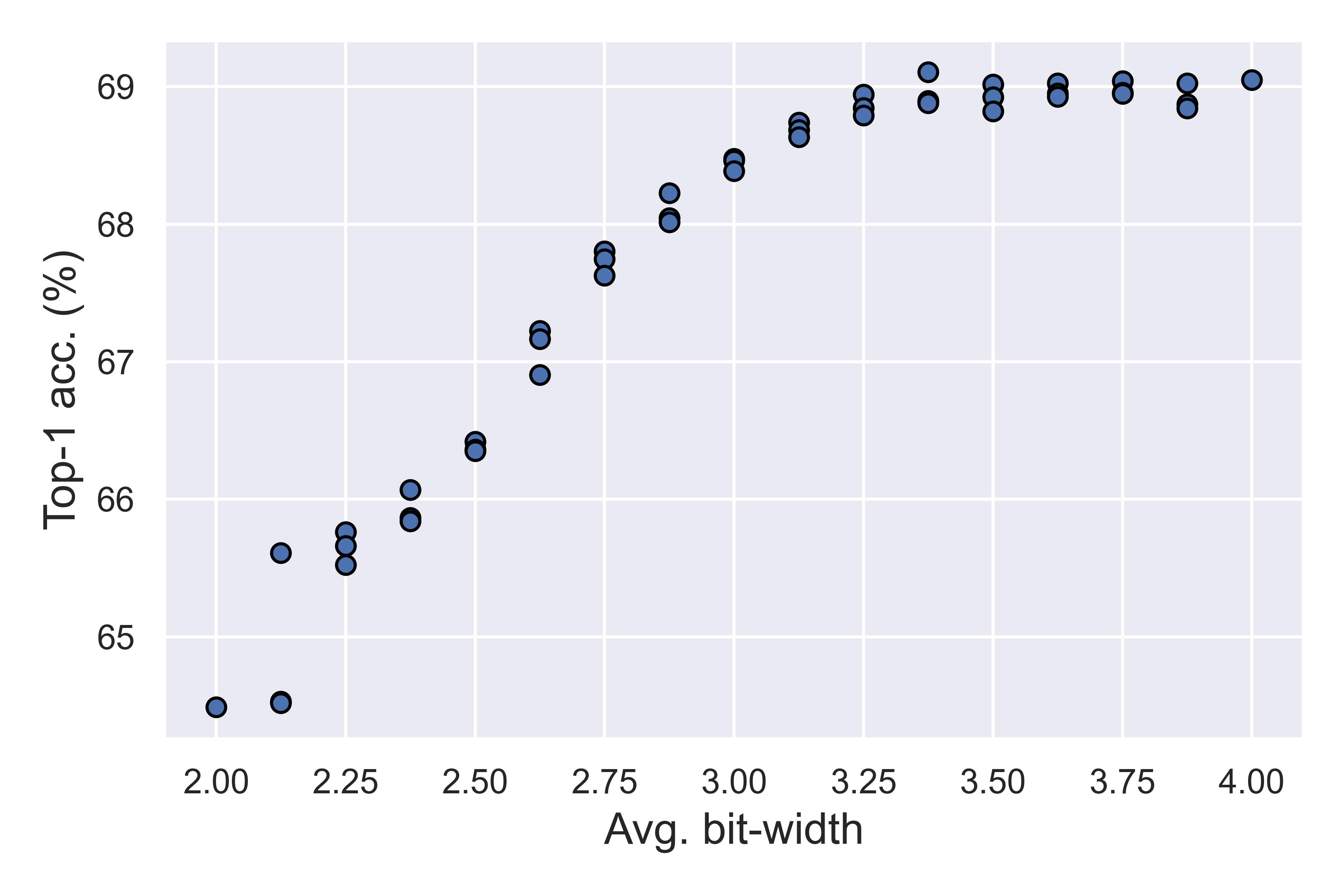}
    \caption{Top-1 accuracy (\%) on ImageNet for a set of  sub-nets extracted from Bit-Mixer's meta-ResNet-18. Note that the accuracy smoothly varies as the avg. bit width changes.}
    \label{fig:sub-netsts}
\end{figure}

Bit-Mixer's trained meta-network contains an exponential number of sub-nets. By changing the bit-width of its individual layers (at runtime, and without extra training), a device or an application where the network is deployed can benefit from a finer trade-off between accuracy and speed. Herein, we describe a method for ``extracting'' highly performing sub-nets from the meta network given a specific avg. bit-width budget. We note that no training is required for finding these sub-nets.

\begin{figure}
    \centering
    \includegraphics[width=7.0cm]{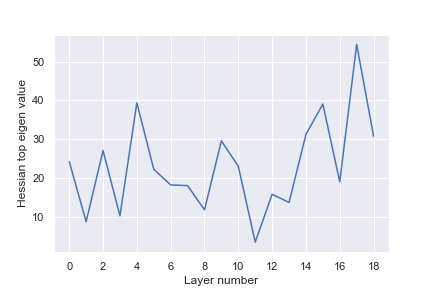}
    \caption{Top eigenvalues of the Hessian matrix for each layer of the network with const. bit-width equal to 4. Notice that layers located towards the end of the network are generally more sensitive to noise. }
    \label{fig:eigen-values}
\end{figure}

To facilitate the selection of interesting (as measured in terms of accuracy per avg. bit-width) candidates out of the given population, we followed~\cite{dong2019hawq}, and for each layer, we computed the top eigenvalue of the Hessian~\footnote{Since forming the entire Hessian matrix is computationally and memory prohibitive, we made use of the power iteration algorithm~\cite{martens2010deep}.}. Note that, for this purpose, we used the network with the highest possible bit-width (i.e. constant bit-width equal to 4 for all layers). The top eigenvalues computed per each layer for the network can be seen in Fig.~\ref{fig:eigen-values}.  In general, smaller eigenvalues correspond to flatter loss surfaces, which in turn, suggest that such layers are good candidates for more aggressive quantization given that the induced errors are less likely to be amplified~\cite{hochreiter1997flat}.  

Given a meta-network $\Phi$, trained as described in Section~\ref{ssec:optimization-strategy}, and a target average bit-width $b_{avg}=\frac{1}{N}\sum_i^N b_i$, where N is the number of layers and $b_i$ the selected bit-width of the $i$-th layer, we attempt to identify a set of promising sub-nets $\{\Phi_0,...,\Phi_m\}$ formed by changing the per-layer bit-width as follows: Let $C_{bits} = \lceil N \times b_{avg}\rfloor$ be the total bit-cost of the desired sub-net, and $\mathbf{v}_C \in \mathbb{R}^{N}$ a per-layer defined cost-vector constructed by taking the highest eigenvalue of the Hessian matrix of each layer. Note that, depending on the target scenario, the cost could be adjusted to take into consideration device-specific knowledge. Since the set of sub-nets $\{\Phi_0,...,\Phi_m\}_C$ of cost C is finite, a straightforward approach is to use a greedy approach generating all potential candidates of bit-cost $C_{bits}$. Once generated, for each configuration from the set, we compute its final ranking cost by taking the product  $C_{total}=[b_0 b_1 ... b_{N}] \times \mathbf{v}_C^T$. We can then select the top-k candidates and evaluate their accuracy. Fig.~\ref{fig:eigen-values} shows a few candidates for various avg. bit-widths alongside their corresponding accuracy. It can be observed that even by using a simple method like the one described above, a diverse set, in terms of accuracy,  of networks can be obtained covering the whole spectrum of avg. bit-widths. 

\subsection{Effect of knowledge distillation}

Knowledge distillation has been previously shown to improve the performance of both full precision~\cite{hinton2015distilling} and quantized neural networks~\cite{mishra2017apprentice,martinez2020training}. Herein, we analyze and validate to what extent distillation helps improve the training of Bit-Mixer for both Stages II and III. In particular, we explore two scenarios for the teacher: (1) Using a full precision network (FP32), and (2) using the trained network after Stage I. We note that, in all cases, the student and teacher networks have exactly the same architecture. As the results from Table~\ref{tab:resnet18_distil} show, distillation does indeed improve the accuracy, although the improvements are lower than typically observed for independently quantized or full precision models. 

\begin{table}[!htbp]
	\caption{Top-1 accuracy (\%) on ImageNet for Bit-Mixer trained with and without distillation.}\label{tab:resnet18_distil}
	\resizebox{\columnwidth}{!}{%
	\centering
		\begin{tabular}{cccccc}
			\toprule

			\multirow{2}{*}{Method} & \multirow{2}{*}{Teacher}  &\multicolumn{4}{c}{Bit-width} \\
			\cline{3-6} 
			& & 4 & 3 & 2 & Rand.   \\			
			\midrule
			\multirow{3}{*}{Ours--Stage II} & - & 69.1 & 68.5 & 65.1 & -  \\
			& Stage I & 69.4 & 68.7 & 65.6 & - \\
			& FP32 & 69.3 & 68.7 & 65.5 & -  \\
            \midrule
			\multirow{3}{*}{Ours--Stage III} & - & 69.0 & 68.4 & 64.0 &  65.5\\
			 & Stage I & 69.1 & 68.6 & 64.5 &  65.7 \\
			 & FP32 &  69.2 & 68.6 & 64.4 & 65.8  \\
			\bottomrule
		\end{tabular}
	}
\end{table}

\begin{figure}[ht]
    \centering
    \begin{subfigure}[t]{0.22\textwidth}
        \includegraphics[width=4.5cm]{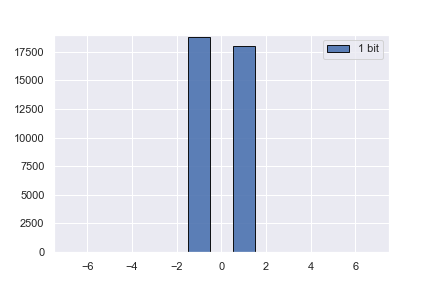}
    \end{subfigure}
    ~
    \begin{subfigure}[t]{0.22\textwidth}
        \includegraphics[width=4.5cm]{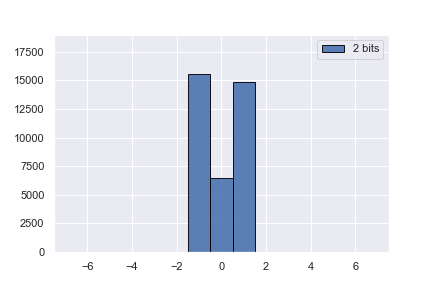}
    \end{subfigure}
    ~
    \begin{subfigure}[t]{0.22\textwidth}
        \includegraphics[width=4.5cm]{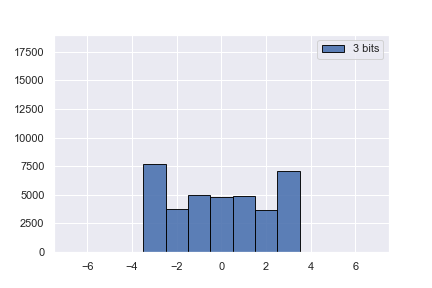}
    \end{subfigure}
    ~
    \begin{subfigure}[t]{0.22\textwidth}
        \includegraphics[width=4.5cm]{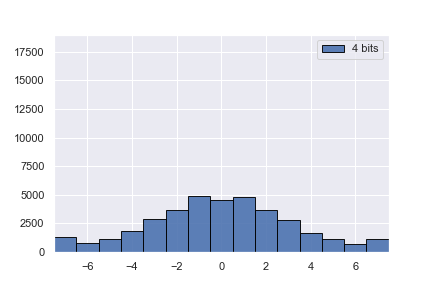}
    \end{subfigure}
    \caption{Weights distribution for 1, 2, 3 and 4 bits after quantization. Note the significant difference in distribution for the 1 bit quantization.}
    \label{fig:weights-distribution}
\end{figure}

\subsection{1--4 bit quantization}
In Section~\ref{sec:method}, we introduced a unified quantization scheme that can be used across all bit-widths, including 1 bit  quantization (\ie binarization). As our results in Table~\ref{tab:resnet18_1bit} suggest, using a single, shared, set of weights, we can successfully train the meta-network up to Stage II using all 4 bit-widths (\ie 4,3,2 and 1) with minimal accuracy loss~\footnote{We note, that unlike the current paradigm used in most recent works on network binarization that is to maintain the $1\times1$ downsampling layers to full precision~\cite{martinez2020training}, in our experiments we binarize them too.}. This is performed in order to align the 1 bit quantization to the rest ones. The training scheduler used for the 4-3-2-1 quantization is the same as the one used for 4-3-2. Notice that despite the relatively larger accuracy gap between binarization and 4 bit quantization, also observed from the noticeably different weight distribution as shown in Fig.~\ref{fig:weights-distribution}, the trained model after Stage II offers overall a good accuracy.

\begin{table}[!htbp]
	\caption{Top-1 accuracy (\%) on ImageNet using a ResNet-18 for 4-3-2-1 bits quantization.}\label{tab:resnet18_1bit}
	\centering
		\begin{tabular}{ccccc}
			\toprule

			\multirow{2}{*}{Method}  &\multicolumn{4}{c}{Bit-width} \\
			\cline{2-5} 
			& 4 & 3 & 2 & 1  \\			
            \midrule
            Independent & 69.1 & 68.5  & 65.1 & 59.0 \\
            \midrule
            Adabits~\cite{jin2020adabits} & 69.2 & 68.5 & 65.1  & - \\
            Ours (Stage II) & 69.4 & 68.7 & 65.6  & - \\
            Ours (Stage II) & 68.7 & 68.0 & 64.2  & 57.3 \\
			\bottomrule
		\end{tabular}
\end{table}

Following Stage II, we continued the training of the above model to obtain the final, Stage III, 4-3-2-1 Bit-Mixer model. However, during this last stage, the training did not converge to the desired outcome. We believe that the main reason for this is the lack of the zeroth state for the binary case. Specifically, while the 2-4 bit-width quantization share the lower states between themselves, as shown in Fig.~\ref{fig:q-ours}, the same is nor true for 1 bit quantization, which lacks the zeroth state, introducing high quantization errors around it and resulting in a different distribution (see Fig.~\ref{fig:weights-distribution}).

However, we did manage to train successfully a Bit-Mixer model with the following configuration $b_{act}=\{2,3,4\}$ for the activations and $b_{w}=\{1,3,4\}$ for the weights, respectively. The results are shown in Table~\ref{tab:resnet18_1w-2a}. Instead of using 2 bits for the activations and weights, in this case, we binarize the later. Since our 2 bit representation is, in fact, a ternary one, this ternary-binary quantization allows for efficient bit-wise implementation too which can result in at least $40\times$~\cite{wan2018tbn} faster convolutions. 

\begin{table}[!htbp]
	\caption{Top-1 accuracy (\%) on ImageNet using a ResNet-18 for 4-3-1.5 bits quantization. * -denotes binary-ternary quantization}\label{tab:resnet18_1w-2a}
	\centering
		\begin{tabular}{ccccc}
			\toprule

			\multirow{2}{*}{Method}  &\multicolumn{4}{c}{Bit-width} \\
			\cline{2-5} 
			& 4 & 3 & 1.5* & rand\\			
            \midrule
            Adabits~\cite{jin2020adabits} & 69.2 & 68.5 & -  & - \\
            \midrule
            Ours - Stage II & 69.0 & 68.7 & 64.0  & - \\
            Ours - Stage III & 69.0 & 68.5 & 62.1  & 62.9 \\
			\bottomrule
		\end{tabular}
\end{table}

\subsection{Scale-- vs. clip--based mixed quantization}

\begin{figure}
    \centering
    \includegraphics[width=7.0cm]{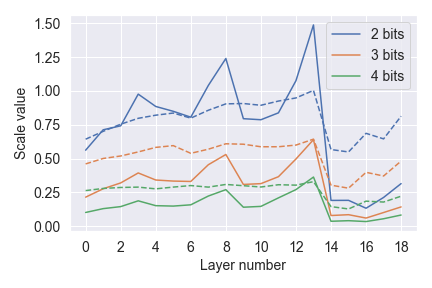}
    \caption{Quantization scales for the activations and weights (dashed line) of each layer of a ResNet-18 model quantized to $b=\{2,3,4\}$. Notice that the ratio between each scale is approximately equal with that of their corresponding maximum representable values.}
    \label{fig:scales}
\end{figure}

Throughout this work, we quantize our models using Eq.~\ref{eq:ours-q} and~\ref{eq:ours-q-op}. Fig.~\ref{fig:scales} shows how the learnable quantization scaling factors $\alpha$ in Eq.~\ref{eq:ours-q-op} change their value as we advance through the network. Importantly, the ratio between $\alpha_i$ and $\alpha_j$ is approximately equal to that of $m_i$/$m_j$ suggesting that all bit-widths are roughly scaled to occupy the whole range. 

To emphasise the importance of the per bit-width scaling factors, we also tested a slightly different approach. The idea is to have shareable quantized weights (and activations) which are obtained by firstly quantizing the real-valued weights to the maximum bit-width $n$ (\ie $-m_n$ and $m_n$) and then clipping them to fit the required bit-width (\ie $-m_i$ and $m_i$). For the weights, this idea is described by:   
\begin{equation}
\begin{split}
\mathbf{\Tilde{W}}_n = \texttt{q'}_n(\texttt{clip}(\frac{W}{\alpha}, -m_n, m_n))
\end{split}
\end{equation}
\begin{equation}
\begin{split}
\mathbf{\Tilde{W}}_i = \alpha \times \texttt{clip}(\mathbf{\Tilde{W}}_{i+1}, -m_i, m_i), \\ 
\end{split}
\end{equation}
Intuitively, this has the effect that the largest (in magnitude) values of the real-valued weights are mapped to states/bits which are present only in the higher bit-widths. On the contrary, using per bit-width scaling factors, the whole range of real-valued weights is mapped to the whole range corresponding to a specific bit-width (\ie $-m_i$ and $m_i$). When trained with this type of quantization, we found that the obtained networks performed 5--10\% worse than with the scale-based quantization.

\subsection{Symmetric vs Asymmetric quantization}

We firstly, note that, in this work, symmetric quantization to refers to the case where the data are mapped to integers in the range $\{-2^{b-1}-1,...,2^{b-1}-1\}$ while asymmetric to the case where the data are mapped into $\{-2^{b-1},...,2^{b-1}-1\}$. In both cases, we consider 0 itself as the zero point as this allows for more efficient implementations. To ensure that no accuracy loss occurs due to the aforementioned design choices, we trained three models: One with asymmetric quantization, one with symmetric, and finally one with symmetric quantization but using the \texttt{round} function in Eq.~\ref{eq:ours-q-op} instead of \texttt{floor}. As the results from Table~\ref{tab:resnet18_quant_schemes} show, all 3 variants are producing essentially identical results.

\begin{table}[!htbp]
	\caption{Top-1 accuracy (\%) on ImageNet using a standard ResNet-18 quantized to 2, 3 and 4 bits using 3 different quantization schemes.}\label{tab:resnet18_quant_schemes}
	\centering
		\begin{tabular}{cccc}
			\toprule

			\multirow{2}{*}{Quantization Method}  &\multicolumn{3}{c}{Bit-width} \\
			\cline{2-4} 
			& 4 & 3 & 2 \\			
            \midrule
            symmetric  & 69.1 & 68.5 & 65.1   \\
            asymmetric  & 69.2 & 68.5 & 65.2   \\
            asymmetric with round & 69.2 & 68.6 & 65.2  \\
			\bottomrule
		\end{tabular}
\end{table}

\begin{table}[ht]
    \centering
    \resizebox{\columnwidth}{!}{%
    \begin{tabular}{c|cccccc}
    \toprule
    Method & 32 & 4 & 3.5 & 3 & 2.5 & 2\\ 
    \midrule
    DoReFa~\cite{zhou2016dorefa} & 70.4 & 68.1 & - & 67.5 & - & 62.6 \\
    LQ-Net~\cite{zhang2018lq} & 70.3 & 69.3 & - & 68.2 & - & 64.9 \\
    PACT~\cite{choi2018pact} & 70.4 & 69.2 & - & 68.1 & - & 64.4 \\
    QIL~\cite{jung2019learning} & 70.2 & 70.1 & - & 69.2 & - & 65.7 \\
    DSQ~\cite{gong2019differentiable} & 69.9 & 69.6 & - & 68.7 & - & 65.2 \\
    APoT~\cite{li2020additive} & 70.2 & - & - & 69.9 & - & - \\
    EdMIPS~\cite{cai2020rethinking}* & - & 68 & 67.7 & 67 & 66.4 & 65.9\\
    Adabits~\cite{jin2020adabits} & - &69,2 & - & 68.5 & - & 65.1\\
    \midrule
    Ours & 69.6 & 69.1 & 69.2 & 68.6 & 66.4 & 64.4\\
    \bottomrule
    \end{tabular}
}
    \caption{Comparison against the state-of-the-art in fixed-bit and mixed precision quantization in terms of top-1 accuracy (\%) on ImageNet using a ResNet-18 architecture. * refers to results where either the number of bits or the accuracy is approximately the one stated in the table.}
    \label{tab:comparison_nas}
\end{table}

\begin{table}[!htbp]
	\caption{Top-1 accuracy (\%) on ImageNet obtained by applying Bit-Mixer on several ResNet and EBN architectures. The accuracy of AdaBits is directly comparable with Ours-Stage II. Notice that Bit-Mixer (Ours) is the only method that can produce a result for layer-wise \textit{rand.} bit allocation. Note that, as shown in Section~\ref{ssec:hawq}, certain sampled sub-nets from our Bit-Mixer meta-network are significantly more accurate than \textit{rand.} * - denotes result taking directly from~\cite{jin2020adabits}.}\label{tab:comparison-sota}
	\centering
	\resizebox{\columnwidth}{!}{%
		\begin{tabular}{cccccc}
			\toprule
            \multirow{3}{*}{Arch.} & \multirow{3}{*}{\#bits} & \multicolumn{4}{c}{Method} \\
			\cline{3-6} 
			& & Indep. & AdaBits & Ours & Ours   \\			
			& & & ~\cite{jin2020adabits} & (Stage II) & \\
			\midrule
            \multirow{4}{*}{ResNet-18} & 4 & 69.1 & 69.2 & 69.4 & 69.2\\
             & 3 & 68.5 & 68.5 & 68.7 & 68.6 \\
             & 2 & 65.1 & 65.1 & 65.6 & 64.4 \\
             & Rand. & - & - & - & 65.8 \\
             \midrule
            \multirow{4}{*}{ResNet-34} & 4 & 73.1 & 73.0 & 73.0 & 72.9 \\
             & 3 & 72.6 & 72.5 & 72.6 & 72.5 \\
             & 2 & 70.2 & 70.0 & 70.1 & 69.6\\
             & Rand. & - & - & - & 70.5\\
             \midrule  
            \multirow{4}{*}{ResNet-50} & 4 & 75.5   & 76.3*  & 75.2 & 75.2\\
             & 3 & 75.3 &  75.9* & 74.9 & 74.8\\
             & 2 & 72.8  & 73.3* & 72.7 & 72.1\\
             & Rand. & - & - & - & 73.2\\
             \midrule   
            \multirow{3}{*}{EBN} & 4 & 74.0 & - & 74.0 & 73.9\\
             & 3 & 73.5 & - & 73.4 & 73.3\\
             4:8:8:16 & 2 & 70.7 & - & 70.5 & 70.4\\
             & Rand. & - & - & - & 71.8\\
             \midrule   
            \multirow{3}{*}{EBN} & 4 & 73.8 & - & 73.8 & 73.3\\
             & 3 & 73.3 & - & 73.2 & 72.8\\
             4:8:16:32 & 2 & 69.8 & - & 69.7 & 68.9\\
             & Rand. & - & - & - & 70.0\\
             \midrule
            \multirow{3}{*}{EBN} & 4 & 74.7 & - & 74.6 & 74.7\\
             & 3 & 74.2 & - & 74.2 & 74.2\\
             4:4:4:4 & 2 & 71.5 & - & 71.1 & 71.4\\
             & Rand. & - & - & - & 72.1\\
			\bottomrule
		\end{tabular}
		}
\end{table}

\section{Results}

\subsection{Experimental setup}~\label{ssec:exp_setup}

All out experiments are performed on ImageNet~\cite{deng2009imagenet}. We focus on the 2-4 bits quantization range. For $b>4$, the accuracy almost always matches or gets very close to that of the full precision counter-parts. Following previous work (e.g. ~\cite{jin2020adabits,rastegari2016xnor,cai2020rethinking,zhou2016dorefa} ) the batch normalization layers are not quantized.

\noindent\textbf{Network architectures:} In order to cover a broad spectrum of architectures in terms of depth, width and cardinality (\ie via grouped convolutions) we performed experiments using the following architectures: (a) ResNet~\cite{he2016deep} (18, 34, 50) and, (b) the recently proposed EBN of~\cite{bulat2020high}. We chose the later since it was  shown to be efficient, suitable for quantization and flexible in terms of varying the width and the group size of the model easily. For example, by increasing the group size, more efficient variants can be obtained. Note that we did not use expert convolutions as proposed in~\cite{bulat2020high}. We used an EBN which, similarly to a Resnet-18, has 4 stages and 2 convolutional blocks per stage. The width of each stage is double the one used in Resnet-18. Finally, the group size per stage is denoted by G0:G1:G2:G3. We tried 3 EBN variants in total: 4:8:8:16, 4:8:16:32 and 4:4:4:4.

\noindent\textbf{Training details:} Unless otherwise stated, all models are trained following the same recipe: the networks are trained for 160 epochs using a cosine scheduler with warm-up (10 epochs) and no restarts~\cite{loshchilov2016sgdr} with a starting learning rate of $0.001$ and a weight decay of $1e-4$. We used the Adam optimizer~\cite{kingma2014adam}. For augmentation, we follow the standard set of transformations used for ImageNet in prior works, mainly: random crop, resize to $224\times224$px and random flipping. For stage III, we gradually increase the probability of $1-\sigma$ from 0 to a target value $k$ during early training, until the network configurations stabilize. For the rest of the training (typically after epoch 80), $k$ remains fixed. The value of $k$ is determined based on the network architecture (typically $2/3<k<4/5$). During evaluation, we resize the images to $256\times256$px and then center crop them to the same $224\times224$px resolution. All experiments are implemented using PyTorch~\cite{paszke2019pytorch}.

\subsection{Comparison with state-of-the-art}

Herein, we firstly compare our method against the current state-of-the-art in quantization. We note that it is hard to make a direct comparison between Bit-Mixer and other methods, as our method: (1) is the very first of its kind that offers the flexibility of layer-wise bid-width selection during runtime, (2) is not focusing on maximizing accuracy for a specific bit-width like other fixed-bit quantization methods, nor (3) is focusing on finding the optimal bit-width allocation for maximizing accuracy like other mixed-precision methods. Moreover, (4) the accuracy results reported in other papers depend on other factors, for example, a very important one is the accuracy of the original FP32 model used. Hence, the main aim of these comparisons is to rather illustrate that the networks trained with Bit-Mixer offer accuracy in par with recently proposed state-of-the-art quantization methods. 

To this end, in Table~\ref{tab:comparison_nas}, we report our results in comparison with a variety of recently proposed state-of-the-art methods for fixed-bit and mixed precision quantization~\cite{zhou2016dorefa, zhang2018lq, choi2018pact, jung2019learning, gong2019differentiable, li2020additive, jin2020adabits, cai2020rethinking}. In all cases, the ResNet-18 architecture was used. As it can be observed, Bit-Mixer provides very competitive results by just training a \textit{single} meta-network which can dynamically define the per-layer bit-width at runtime. This is very important as our goal is to have the flexibility that Bit-Mixer can offer without however compromising the capacity for highly accurate inference. 

This section, and, in particular, Table~\ref{tab:comparison-sota}, also provides results obtained by training Bit-Mixer meta-networks using the ResNet and EBN architectures detailed in Section~\ref{ssec:exp_setup}. Where possible, we also compare with Adabits~\cite{jin2020adabits}. Note that Bit-Mixer after Stage II (Ours-Stage II) is directly comparable with Adabits. Note also that Bit-Mixer after Stage III (Ours) is the only method that can provide layer-wise \textit{random} bit allocation. We believe that the results of Table~\ref{tab:comparison-sota} conclusively show that Bit-Mixer can be successfully applied to train meta-networks across a wide variety of network architectures.

\section{Conclusions}

To the best of our knowledge, this work constitutes the very first attempt to training a meta-network with shared weights the layers/blocks of which can be independently quantized to any desired bit-width at runtime. To this end, we made two key contributions: (a) Transitional Batch-Norms and (b) a 3-stage optimization pipeline which is shown capable of training such a network. We presented a series of ablation studies analyzing important components and features of the proposed method. Moreover we presented comparisons with several state-of-the-art quantization methods as well as results obtained by applying Bit-Mixer on several architectures. These results show that our method can successfully train a meta-network with arbitrary layer-wise bit-width selection without compromising accuracy.

{\small
\bibliographystyle{ieee_fullname}
\bibliography{egbib}
}

\end{document}